\documentclass[11pt]{article}

\usepackage{acl}

\usepackage{times}
\usepackage{latexsym}

\usepackage[T1]{fontenc}

\usepackage{microtype}

\usepackage{inconsolata}

\usepackage{microtype}

\usepackage{inconsolata}
\usepackage[textsize=tiny]{todonotes}

\usepackage{graphicx}
\usepackage{multirow}
\usepackage{multicol}
\usepackage{tabularx}
\usepackage{amssymb}
\usepackage{mathtools}
\usepackage{enumitem}
\usepackage{amsthm}
\usepackage{bbm}
\usepackage{comment}
\usepackage{caption}
\usepackage{subcaption}
\usepackage{wrapfig}
\usepackage{pifont}
\usepackage{tabularx}
\usepackage{booktabs}
\usepackage{colortbl}
\usepackage{url}
\usepackage{algorithm}
\usepackage{algpseudocode}

\usepackage{booktabs}
\usepackage{amsmath}
\usepackage{subcaption}
\usepackage{multirow}
\usepackage{soul}
\usepackage{color, colortbl}
\usepackage{wrapfig}

\usepackage{lipsum}

\newcommand{\hlc}[2][yellow]{{%
    \colorlet{foo}{#1}%
    \sethlcolor{foo}\hl{#2}}%
}

\usepackage[skins,breakable]{tcolorbox}

\title{ReAttn: Improving Attention-based Re-ranking via Attention Re-weighting}

\author{Yuxing Tian$^1$, Fengran Mo$^1$, \textbf{Weixu Zhang}$^{2}$, \textbf{Yiyan Qi}$^3$, \textbf{Jian-Yun Nie}$^1$\\
$^1$University of Montreal; 
$^2$McGill University \& MILA; \\
$^3$International Digital Economy Academy \\
}

\begin{document}
\maketitle

\begin{abstract}
The strong capabilities of recent Large Language Models (LLMs) have made them highly effective for  zero-shot re-ranking task. Attention-based re-ranking methods, which derive relevance scores directly from attention weights, offer an efficient and interpretable alternative to generation-based re-ranking methods. However, they still face two major limitations. First, attention signals are highly concentrated a small subset of tokens within a few documents, making others indistinguishable. Second, attention often overemphasizes phrases lexically similar to the query, yielding biased rankings that irrelevant documents with mere lexical resemblance are regarded as relevant. In this paper, we propose \textbf{ReAttn}, a post-hoc re-weighting strategy for attention-based re-ranking methods. It first compute the cross-document IDF weighting to down-weight attention on query-overlapping tokens that frequently appear across the candidate documents, reducing lexical bias and emphasizing distinctive terms. It then employs entropy-based regularization to mitigate over-concentrated attention, encouraging a more balanced distribution across informative tokens. Both adjustments operate directly on existing attention weights without additional training or supervision. Extensive experiments demonstrate the effectiveness of our method.
\end{abstract}

\section{Introduction}

\begin{figure}[t]
    \centering
    \includegraphics[width=1.0\linewidth]{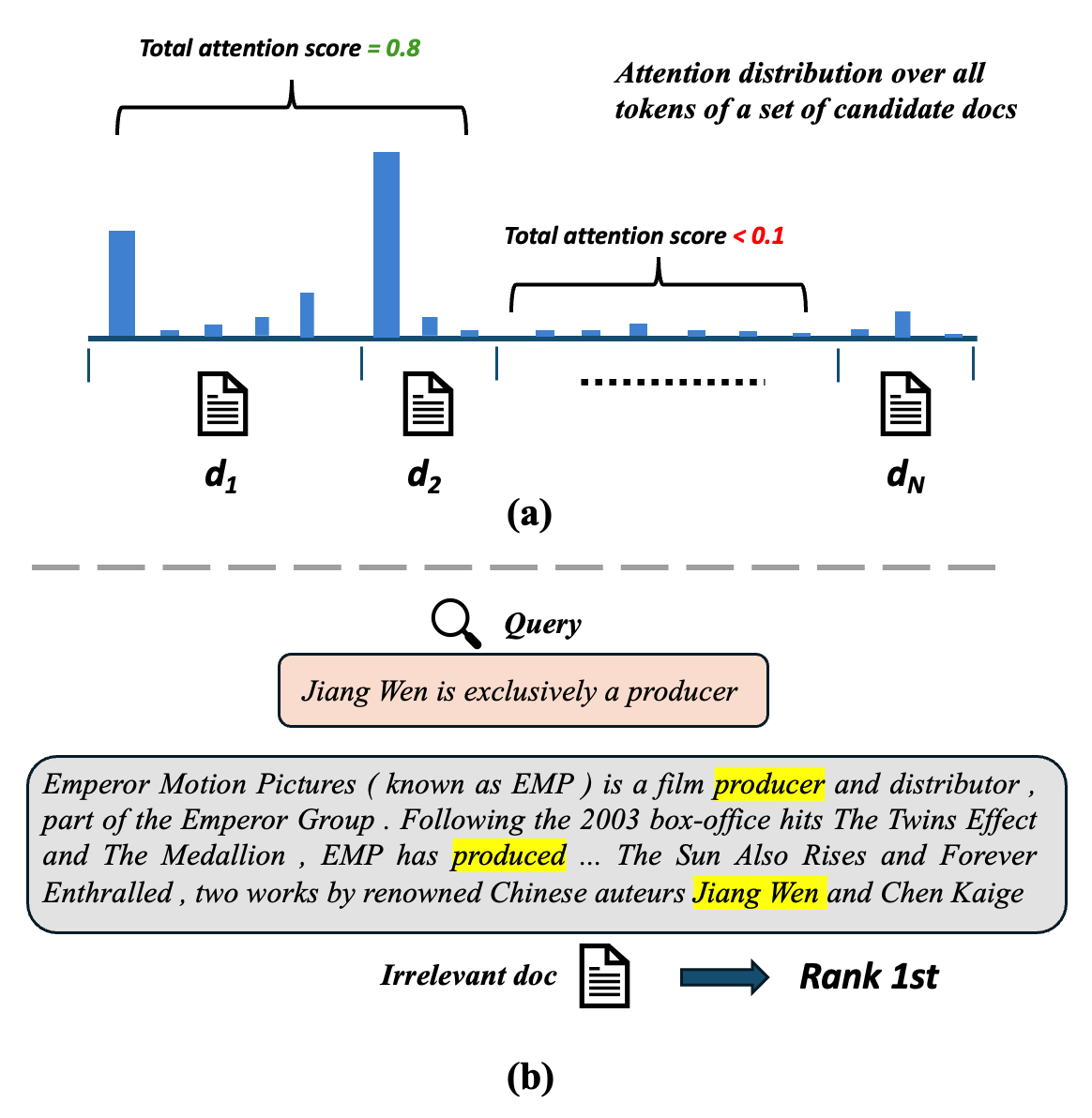}
    \caption{Illustration of two issues in attention-based re-ranking. (a) Signal concentration: the total attention mass is heavily concentrated on a few tokens of a few documents, leaving most candidates with negligible attention scores. (b) Lexical bias: attention disproportionately highlights tokens that are lexically similar to the query (e.g., producer, Jiang Wen), causing irrelevant documents with lexical overlap to the query to receive inflated relevance scores and misleading the ranking.}
    \label{fig:toy_example}
\end{figure}
Information retrieval (IR) serves as a cornerstone of modern intelligent systems, powering applications in search, recommendation, and retrieval-augmented generation~\citep{lewis2020rag}. A typical IR pipeline adopts a two-stage paradigm in which an initial retriever gathers a broad set of candidate documents using sparse~\citep{bm25} or dense retriever~\citep{karpukhin2020dense}, and a subsequent re-ranker refines these candidate documents to produce the final ranked list~\citep{nogueira2019multi, nogueira2020document}. The re-ranking stage is particularly critical, as it governs retrieval precision and shapes the contextual information accessible to downstream reasoning or generation models~\citep{gao2023retrieval}.

Traditional re-ranking methods have predominantly relied on supervised learning with cross-encoder architectures, which require substantial labeled data and task-specific fine-tuning. The advent of large language models (LLMs) has profoundly transformed this landscape. Owing to their strong zero-shot reasoning capabilities and ability to comprehend long contexts, LLMs can effectively perform zero-shot re-ranking task~\citep{sachan2022improving, sun2023chatgpt, qin2024large}. Early attempts to apply LLMs to re-ranking can be viewed as generation-based approaches~\citep{sun2023RankGPT}, as they directly prompt the model to generate an ordered list of candidate documents identifiers, effectively framing re-ranking as a text generation task. Although these approaches exploit the generative capabilities of LLMs to reason about document relevance, they also introduce several practical challenges: high computational overhead from autoregressive decoding; strong sensitivity to prompt design; and low instruction-following reliability, which frequently results in inconsistencies or omissions in the generated outputs. Recent research has increasingly focused on non-generative re-ranking approaches that derive relevance score of documents directly from internal signals within LLMs. ~\citet{chenattention} has shown that the attention signal in LLMs implicitly reflect relevance between queries and documents: query tokens tend to allocate greater attention weights to tokens in relevant documents. Specifically, they first aggregate attention weights that tokens in each document receive from all query tokens across all layers and heads of the LLM to obtain the token-level relevance scores. These scores are then summed within each document to produce a document-level relevance score used for ranking. \citet{zhang2025query} further identify specialized query-focused retrieval heads that attend selectively to information most relevant to the query. Compared with~\citet{chenattention}, which aggregates attention weights across all heads, using only these specialized heads obtains more reliable relevance score and leads to better re-ranking performance.

Despite the promising performance of attention-based re-ranking methods, two inherent issues constrain its effectiveness. The first is signal concentration (Figure~\ref{fig:toy_example}(a)): a small number of tokens within a few documents absorb most of the total attention mass, while most documents receive minimal attention contributions. This imbalance limits the re-ranker’s ability to distinguish and rank the less salient documents. The second is lexical bias (Figure~\ref{fig:toy_example}(b)), where attention tends to overemphasize tokens that are lexically similar to the query. As a consequence, documents containing such overlapping expressions, even when semantically irrelevant, can receive high attention scores. While attention mechanisms are capable of modeling semantic relevance, this lexical preference can partially constrain their effectiveness, leading to suboptimal ranking of truly relevant documents.

To this end, we propose \textbf{ReAttn}, a post-hoc adjustment strategy for attention-based re-ranking that re-weighting token- and document-level attention score without additional training. We first introduce a cross-document inverse document frequency (IDF) weighting to mitigate lexical bias. Intuitively, if a query token appears across most documents, it provides little evidence for distinguishing truly relevant documents and dilutes the discriminative strength of the attention signal. Thus, we measure the frequency of each query token appear across the candidate set and compute corresponding cross-document IDF weights. These weights are then used to rescale token-level attention scores, reducing the influence of ubiquitous tokens and emphasizing distinctive ones that better indicate relevance. Next, we introduce an entropy-based regularization. For each document, we normalize its token-level relevance scores into a probability distribution and compute its entropy as a measure of attention dispersion. Documents with higher entropy reflect broader coverage across informative tokens, whereas lower entropy signals an overly narrow focus. Then we apply a document-level weighting based on this measure, decreasing the relevance scores of low entropy documents and increasing those of high entropy documents. This yields a more balanced allocation of attention across documents and improves discrimination among moderately relevant documents.

Our contributions are summarized as follows:
\begin{itemize}
    \item We identify two core issues that limit the effectiveness of attention-based re-ranking methods: signal concentration and lexical bias.
    \item We propose ReAttn, which introduces a cross-document IDF weighting to reduce lexical bias and an entropy-based regularization to alleviate score concentration, jointly improving attention reliability and ranking stability.
    \item Extensive experiments on multiple datasets of re-ranking and long context reasoning task demonstrate that ReAttn consistently improves ranking performance over attention-based re-ranking baselines with minimal computational overhead.
\end{itemize}

\section{Related Work}

\textbf{Zero-Shot Re-Ranking with LLMs.} Zero-shot re-ranking with LLMs broadly categorized into three primary paradigms: point-wise, pair-wise, and list-wise approaches.
In point-wise re-ranking, each candidate document is evaluated independently, typically using generation probabilities or scalar relevance scores derived from the LLM's logits~\citep{sachan2022improving, liang2022holistic}. Although computationally efficient, these methods often fall short in performance because they fail to capture interactions or relative preferences among documents.
Pair-wise methods \citep{qin2024large} address this limitation by comparing document pairs to infer preference relations, which are then aggregated to produce the final ranking. While this pairwise comparison generally enhances discriminative accuracy, it also introduces a quadratic computational cost with respect to the number of candidate documents.
In contrast, list-wise approaches \citep{ma2023zero, sun2023chatgpt, chenattention} process the entire candidate set simultaneously, enabling the model to capture holistic dependencies and optimize ranking quality at the list level. The main challenge of list-wise modeling lies in its requirement for extended context handling. However, recent advances in long-context LLM architectures \citep{chen2023extending, jin2024llm,fu2024data,mo2025uniconv} have made such modeling increasingly practical. As a result, list-wise re-ranking has emerged as both a scalable and highly effective strategy.
Building on this trend, our work focuses on attention-based list-wise re-ranking \citep{chenattention}, which combines the efficiency of non-generative architectures with the stability of attention-driven scoring, outperforming generation-based re-ranking methods in both computational and empirical robustness~\citep{ma2023zero, sun2023chatgpt,mo2023convgqr}.

\textbf{Attention Mechanisms.} Attention mechanisms form the foundation of transformer-based language models, facilitating token-to-token information exchange and offering a potential lens for interpreting model behavior. Since the seminal work of \citet{bahdanau2015attention}, attention weights have often been employed as proxies for token-level importance, providing insights into how models distribute focus during inference. Despite ongoing debate regarding their faithfulness as explanations of internal reasoning \citep{serrano-smith-2019-attention, wiegreffe-pinter-2019-attention,zhang2024blind,zhang2025entropy,zhang2025ratt,mo2026opendecoder}, attention patterns have demonstrated considerable utility in practice. For example, \citet{izacard2021distilling} exploit cross-attention between queries and passages to enhance retriever training, while \citet{Peysakhovich2023AttentionSC} leverage attention-based document sorting to improve text generation quality. Building on this intuition, in-context re-ranking (ICR) \citep{chenattention} interprets attention as an implicit signal of document relevance, enabling LLMs to reorder retrieved passages directly within their context windows, without the need for additional supervision or parameter updates.

Concurrently, research in mechanistic interpretability has examined the functional specialization and redundancy of individual attention heads in transformers. Early investigations demonstrated that many attention heads contribute little to overall performance, with only a small subset being functionally critical \citep{michel2019sixteen, voita2019analyzing}. Subsequent analyses have identified distinct behavioral roles, such as induction heads that capture recurring patterns across sequences \citep{olsson2022context, yinattention, ren2024identifying}, and others that help manage knowledge conflicts \citep{shi2024ircan, jin2024cutting} or mitigate contextual distractions \citep{zhu2025focus}. More recently, attention-based retrieval behavior has emerged as a focal point of study: \citet{wu2025retrieval} identify retrieval heads that copy relevant answer tokens from long contexts, while \citet{zhang2025query} introduce query-focused retrieval heads (QR heads) that model fine-grained interactions between queries and supporting evidence.

% In this work, we build on the notion of retrieval heads \citep{wu2025retrieval, zhang2025query} as the mechanisms responsible for retrieving relevant information from long-context inputs -- directly aligning with the re-ranking objective.

% However, we aggregate the attention weights over all query tokens, rather than only considering the last token as in previous work. We also propose a calibration method to help mitigate LLMs' intrinsic biases. Our experiments on information retrieval benchmarks show the necessity of the proposed components.

\section{Preliminary: Attention-Based Re-ranking with LLMs}
\label{sec:pre}

Formally, let a query be denoted as $Q$ and its corresponding set of retrieved candidate  documents as $D = \{ d_1, d_2, \dots, d_N \}$.  The objective of re-ranking is to assign each document $d_i\in D$ a relevance score with respect to the query $Q$ and a final ranking sorted in descending order of relevance score. In attention-based re-ranking, we first construct the sequence $X=[d_1,d_2,...,d_N,Q]$ by concatenating all candidate documents followed by the query, and provide it as input to a LLM with $L$ transformer layers and $H$ attention heads. For each document $d_i$, the token-level relevance score of its $j$-th token, denoted as  $s_{d_{i,j}, Q}$, is obtained by aggregating the attention weights that this token receives from all query tokens across all layers and attention heads:
\begin{equation}
  s_{d_{i,j}, Q} =\frac{1}{|\mathcal{I}_Q|} \sum_{l=1}^{L} \sum_{h=1}^{H} \sum_{k\in\mathcal{I}_{Q}} a^{l,h}_{j,k}
\end{equation}
where $\mathcal{I}_{Q}$ denotes the set of query tokens, and $a^{l,h}_{j,k}$ represents the attention weight from the $k$-th token (in the query) to the $j$-th token (in document $d_i$) by the $h$-th attention head at $l$-th layer. Summing token-level scores within a document yields the relevance score for document $d_i$.

However, raw attention do not always provide a faithful measure of relevance. Prior work has shown that LLMs exhibit intrinsic biases in attention allocation~ \citep{gallegos2024bias},including disproportionate weighting toward longer documents and meaningless tokens (e.g., punctuation). These intrinsic biases can undermine the reliability of attention-based relevance computation, especially when attention is directly used as a ranking signal. Following \citet{zhao2021calibrate}, ICR~\cite{chenattention} use \textit{``N/A''} as the calibration query $Q_{c}$ and calculate calibration scores $\{s_{{d_{i,j}},Q_{cal}}\}$ via attention aggregation for each document. The calibration scores capture strong attention weights biased towards meaningless tokens in the documents, such as punctuation, which should not affect the relevance of the documents.  Then we can subtract it from ranking scores from the actual query $\{{s_{d_{i,j}, Q}}\}$ to obtain the calibrated ranking score $\{s_{d_{i,j}}\}$:
\begin{equation}
  s_{d_{i,j}} = s_{d_{i,j}, Q} - s_{d_{i,j}, Q_{c}}.
\end{equation}
After that, we filter out such tokens with abnormally negative calibrated scores.
Finally, we sum the calibrated scores for all tokens in each document to obtain the final ranking score $s_{d_i}$, which measures the change of attention weights that each document receives when the query changes from the content-free calibration query to the actual query:
\begin{equation}
    \begin{aligned}
        \mathcal{S}_{d_i} &= \{d_{i,j}\}\\
        \mathcal{S}^*_{d_i} &= \{d_{i,j} | d_{i,j}> \bar{\mathcal{S}}_{d_i} -2\sigma_{\mathcal{S}_{d_i}}\}\\
        s_{d_i} &= \sum_{s\in\mathcal{S}^*_{d_i}}s
    \end{aligned}
\end{equation}
where $\sigma$ denotes standard deviation.

\section{ReAttn}

We propose \textbf{ReAttn}, a post-hoc re-weighting  strategy for attention-based re-ranking that re-weights both token-level and document-level attention scores without any additional training. ReAttn aims to address two key issues: (\textit{i}) \emph{lexical  bias}, where tokens overlapping with the query receive disproportionate attention, and (\textit{ii}) \emph{signal concentration}, where only top-ranked documents receive strong signals, leaving lower-ranked documents poorly differentiated. The method consists of two sequential steps: cross-document IDF reweighting at the token level and entropy-based regularization at the document level. We detail each component below.

\subsection{Cross-Document IDF Weighting}

The first step mitigates lexical bias by down-weighting contributions from query tokens that frequently appear across the candidate set.  Intuitively, if a query token occurs frequently across candidate documents, it provides limited discriminative information and can weaken the attention-based relevance estimation. 

Formally, for each token $t_{i,j}$ in document $d_i$ that also appears in the query $Q$ (i.e., $t_{i,j} \in \mathcal{I}_Q$), we compute its document frequency over the candidate document set $D = \{d_1, \dots, d_N\}$:
\begin{equation}
    \mathrm{df}(t_{i,j}) = \big| \{\, d_k \in D \mid t_{i,j} \in d_k \,\} \big| .
\end{equation}
We then define a normalized inverse document frequency (IDF) weight:
\begin{equation}
    w(t_{i,j}) = \frac{\log \frac{N + 1}{\mathrm{df}(t_{i,j}) + 1}}{\log (N + 1)}
\end{equation}
where $N$ is the total number of documents. Tokens that appear in many documents receive smaller $w(t_{i,j})$, while distinctive tokens retain larger weights. Then the token-level calibrated attention scores $s_{d_{i,j}}$ (from Section~\ref{sec:pre}) are adjusted as:
\begin{equation}
    \tilde{s}_{d_{i,j}} = 
    \begin{cases}
        w(t_{i,j}) \cdot s_{d_{i,j}} & \text{if } t_{i,j} \in \mathcal{I}_Q \\
        s_{d_{i,j}} & \text{otherwise.}
    \end{cases}
\end{equation}
This re-weighting reduces the dominance of ubiquitous query tokens while preserving the influence of distinctive terms.  The document-level base score after IDF re-weighting is obtained by summing token scores within the filtered token set $\mathcal{S}^*_{d_i}$:
\begin{equation}
    B_i = \sum_{j \in \mathcal{S}^*_{d_i}} \tilde{s}_{d_{i,j}} .
\end{equation}
$B_i$ serves as the IDF-adjusted relevance score for document $d_i$, emphasizing tokens that provide stronger discriminative evidence of relevance.

\subsection{Entropy-Based Regularization}
Although IDF weighting alleviates lexical bias, it does not prevent signal concentration, where a few tokens within a few documents dominate the relevance estimation and most documents receive negligible scores. To improve the score dispersion across documents, we introduce an entropy-based regularization that quantifies how evenly attention is distributed within each document. Specifically, for each document $d_i$, we normalize its token-level scores into a probability distribution:
\begin{equation}
    p_{i,j} = \frac{\tilde{s}_{d_{i,j}}}{B_i}, \quad \text{where } \sum_{j \in \mathcal{S}^*_{d_i}} p_{i,j} = 1 .
\end{equation}
The normalized Shannon entropy of this distribution measures the internal dispersion of attention:
\begin{equation}
    E_i = - \frac{\sum_{j \in \mathcal{S}^*_{d_i}} p_{i,j} \log p_{i,j}}{\log |\mathcal{S}^*_{d_i}|}
\end{equation}
where $E_i \in [0,1]$. High-entropy documents have attention spread across multiple informative tokens, while low-entropy documents exhibit overly narrow attention focus. We compute the base-score-weighted mean entropy across all documents:
\begin{equation}
    \bar{E}_B = \frac{\sum_{k=1}^N B_k E_k}{\sum_{k=1}^N B_k} .
\end{equation}
Each document is then assigned a dispersion weight according to its deviation from the average entropy:
\begin{equation}
    W_i = 1 + (E_i - \bar{E}_B) .
\end{equation}
Documents with broader attention coverage ($E_i > \bar{E}_B$) receive a small positive adjustment, whereas narrowly focused ones ($E_i < \bar{E}_B$) are slightly penalized. The adjusted document-level score is:
\begin{equation}
    s'_i = B_i \cdot W_i .
\end{equation}
Finally, to maintain comparability across documents, we normalize the scores so that they sum to one:
\begin{equation}
    s^{\mathrm{final}}_{d_i} = \frac{s'_i}{\sum_{k=1}^{N} s'_k} .
\end{equation}
This entropy-based regularization ensures that attention-based re-ranking captures both strong token-level relevance and balanced document-level coverage. Combined with IDF re-weighting, ReAttn effectively mitigates lexical bias and enhances score discrimination, yielding more robust and fine-grained re-ranking performance.

\begin{table*}[t]
    \centering
        \renewcommand{\tabcolsep}{1.2mm}
        \renewcommand{\arraystretch}{1.2}
 
    \fontsize{8}{8}\selectfont
    % \footnotesize
  
    \begin{tabular}{lcccccccccccc}
    \toprule
& \bf NQ & \bf COVID & \bf NFCorpus & \bf FiQA & \bf Scifact & \bf Scidocs & \bf FEVER & \bf Climate & \bf DBPedia & \bf Robust04 & \bf News & \bf Avg \\
\midrule
BM25 & 30.5 & 59.5 & 32.2 & 23.6 & 67.9 & 14.9 & 65.1 & 16.5 & 31.8 & 40.7 & 39.5 & 38.4 \\
\midrule
& \multicolumn{11}{c}{\textit{Base LLM: Llama-3.2-3B-Instruct }}  \\
\cmidrule{0-12}
RankGPT\textsuperscript{w/o} & 30.0 & 59.5 & 32.2 & 23.6 & 67.9 & 14.9 & 65.9 & 17.1 & 31.8 & 40.7 & 39.5 & 38.5 \\
RankGPT\textsuperscript{Bubble} & 33.2 & 61.8 & 32.0 & 22.4 & 66.1 & 14.8 & 65.8 & 17.1 & 34.8 & 40.5 & 40.2 & 39.0 \\

ICR & 49.2 & 72.3 & 33.8 & 31.8 & 73.3 & 17.4 & 82.6 & 24.2 & 34.7 & 47.2 & 44.7 & 46.5 \\
ICR+ReAttn & 49.8 & 72.5 & 34.4 & 32.5 & 74.1 & 17.5 & 83.0 & 24.5 & 36.8 & 47.7 & 45.4 & 47.3 \\
{QRhead} &  54.9 &  77.4 &  35.1 & 35.1 &  74.7 &  18.3 &  83.7 &  24.5 & 36 & 49.7 & 45.1 &  48.6 \\
QRhead+ReAttn & \bf 55.6 & \bf 77.6 & \bf 36.6 & \bf 36.2 & \bf 75.3 & \bf 19.5 & \bf 84.3 & \bf 25.6 &\bf  38 &\bf  49.8 & \bf 46 & \bf 49.3 \\

\midrule
& \multicolumn{11}{c}{\textit{Base LLM: Llama-3.1-8B-Instruct }}  \\
\cmidrule{0-12}

RankGPT\textsuperscript{w/o} & 30.0 & 59.5 & 32.2 & 23.6 & 67.9 & 14.9 & 65.9 & 16.8 & 31.8 & 40.7 & 39.5 & 38.4 \\
RankGPT\textsuperscript{Bubble} & 53.7 & 75.5 & 34.3 & 31.4 & 69.3 & 17.4 & 67.5 & 23.8 &  
\bf42.9 & 47.8 & 46.2 & 46.3 \\

ICR & 54.0 & 73.3 & 34.8 & 35.6 & 75.5 & 19.0 & 85.8 & 24.8 & 36.9 & 49.0 & 44.5 & 48.5 \\

ICR+ReAttn & 54.4 & 73.5 & 35 & 35.9 & 76.2 & 19.3 & 86.1 & \bf 25.7 & 37.5 & 49.3 & 45.1 & 49.3 \\

{QRhead} & 58.6 & 77.5 & 35.3 & 39.1 & 76.2 &  19.4 & 85.3 & 23.9 & 37.2 &  51.4 &  46.2 & 50.0 \\

QRhead+ReAttn & \bf 59.3 & \bf 78.2 & \bf 36.0 & \bf 40.4 &\bf  76.8 & \bf 20.2 & \bf86.4 & 25.3 & 38.6 & \bf 52.5 & \bf 47 & \bf 50.9 \\
\midrule
 
& \multicolumn{11}{c}{\textit{Base LLM: Qwen-2.5-7B-Instruct }} \\
\cmidrule{0-12}
RankGPT\textsuperscript{w/o} & 30.0 & 59.5 & 32.2 & 23.6 & 67.9 & 14.9 & 65.9 & 16.8 & 31.8 & 40.7 & 39.5 & 38.4 \\
RankGPT\textsuperscript{Bubble} & 42.7 & \bf 70.5 & 34.1 &  29.5 & 69.3 & 16.6 & 70.5 & 19.7 & 37.1 & \bf 46.4 & \bf 43.6 & 43.6 \\

ICR & 43.1 & 66.1 & 32.7 & 27.0 & 71.1 & 16.4 & 79.2 & 19.6 & 35.3 & 43.0 & 40.0 & 43.0 \\
ICR+ReAttn & 44.0 & 67.0 & 33.9 & 28.1 & \bf 72.2 & \bf 17.8 & 80.0 & 20.9 & 36.6 & 43.6 & 41.2 & 44.0 \\
{QRhead} & 49.9 & 67.7 & 33.1 & 29.2 & 71 & 15.3 &  80.7 &  20.1 & 35.7 & 43.7 & 39.8 &  44.2 \\

QRhead+ReAttn & \bf 50.7 & 68.6 & \bf 34.2 &\bf  30.3 & 71.9 & 16.6 & \bf 81.5 & \bf 21.3 & \bf37.4 & 44.8 & 41.0 & \bf 45.3 \\

\midrule

& \multicolumn{11}{c}{\textit{Traditional retrievers}} & \\
\cmidrule{0-12}

Contriever & 44.6 & 67.5 & 32.8 & 28.4 & 67.1 & 18.9 & 64.2 & 28.0 & 39.5 & 45.7 & 41.7 & 43.5 \\

GTR-T5-base & 51.4 & 74.8 & 32.5 & 34.7 & 62.1 & 15.8 & 72.9 & 26.8 & 37.1 & 46.1 & 42.8 & 45.2 \\

BGE-Reranker-base & 55.2 & 66.4 & 31.0 & 31.7 & 70.8 & 15.7 & 88.6 & 36.5 & 42.5 & 39.9 & 37.0 & 46.8 \\

MSMARCO-MiniLM & 55.8 & 74.3 & 35.2 & 35.1 & 68.5 & 17.5 & 80.4 & 25.5 & 45.3 & 47.9 & 43.0 & 48.0 \\
\bottomrule
    \end{tabular}
    \caption{Performance comparison across different base LLMs and re-ranking methods on BEIR benchmarks evaluated by nDCG@10. \textbf{Bold} indicates the best results.}
    \label{tab:beir_main}
\end{table*}

\begin{table*}[t]
    \centering
    \renewcommand{\tabcolsep}{1.2mm}
    \renewcommand{\arraystretch}{1.1}
    % \fontsize{7.75}{7.75}\selectfont
    \small
    \begin{tabular}{lcccccccccccccc}
    \toprule
    && \multicolumn{5}{c}{\textbf{LongMemEval}} && \multicolumn{5}{c}{\textbf{Clipper}} \\
     \midrule
    \multirow{3}{*}{\textbf{\textsc{Retriever}}} && \multicolumn{2}{c}{\textbf{\textsc{Retrieval}}}  && \multicolumn{2}{c}{\textbf{\textsc{End-to-End}}} && \multicolumn{2}{c}{\textbf{\textsc{Retrieval}}}  && \multicolumn{2}{c}{\textbf{\textsc{End-to-End}}} \\
     && \multicolumn{2}{c}{\textbf{\textsc{Recall@k}}}  && \multicolumn{2}{c}{\textbf{\textsc{Performance}}} && \multicolumn{2}{c}{\textbf{\textsc{Recall@k}}}  && \multicolumn{2}{c}{\textbf{\textsc{Performance}}} \\
        
         && k = 5 & k = 10 && Top-5 & Top-10 && k = 3 & k = 5 && Top-3 & Top-5 \\
       \midrule
         \textbf{\textit{Base LLM: Llama-3.2-3B-Instruct}} \\
        Full context && - & - && \multicolumn{2}{c}{28.1} && - & - && \multicolumn{2}{c}{25.2} \\
         BM25 && 57.5 & 67.5 && 46.1 & 44.9 && 74.6 & 83.7 && 20.0 & 22.8 \\
         Contriever && 62.7 & 79.2 && \bf 48.6 & 46.5 && 60.2 & 78.9 && 12.6 & 18.4 \\
         Stella && 63.9 & 77.6 && 44.9 & 47.7 && 83.3 & 90.0 && 21.3 & 25.1 \\
          RankGPT\textsuperscript{w/o} && 1.8 & 3.4  && 23.5 & 23.3 && 16.8 & 27.3 && 3.6 & 8.8 \\
         RankGPT\textsuperscript{Bubble}  && 2.1 & 3.8 && 24.0 & 24.4 && 17.0 & 27.4 && 3.8 & 8.8 \\
         ICR  && 68.7 & 78.8 && 46.5 & 45.1 && 72.8 & 83.6 && 19.4 & 23.6 \\

         ICR+ReAttn  && 68.8 & 79.2 && 46.6 & 45.5 && 72.9 & 83.8 && 19.4 & 23.8 \\
         
         QRhead&&  77.6 &  86.6 && 47.4 & 47.7 &&  85.5 &  93.4 && 23.4 &  26.9 \\
         
         QRhead+ReAttn&& \bf 77.9 & \bf 87.3 && 47.6 & \bf 48 && \bf 85.7 & \bf 93.5 && \bf23.4 & \bf 27.2 \\
         \midrule
         
        \textbf{\textit{Base LLM: Llama-3.1-8B-Instruct}} \\
        Full context && - & - &&  \multicolumn{2}{c}{46.5} && - & - && \multicolumn{2}{c}{31.3} \\
         BM25 && 57.5 & 67.5 && 48.8 & 50.9 && 74.6 & 83.7 && 37.9 & 37.9 \\
         Contriever && 62.7 & 79.2 && 52.6 & 55.4 && 60.2 & 78.9 && 28.2 & 31.1 \\
         Stella && 63.9 & 77.6 && 50.9 & 58.4 && 83.3 & 90.0 && 38.8 & 39.6 \\
        RankGPT\textsuperscript{w/o} && 2.1 & 4.0 && 26.7 & 24.2 && 30.0 & 39.4 && 15.9 & 19.4 \\
        RankGPT\textsuperscript{Bubble} && 8.3 & 9.0 && 28.1 & 27.0 && 36.7 & 44.3 && 19.7 & 20.4 \\
        ICR && 77.0 & 84.4 && 59.3 & 56.1 && 89.3 & 94.7 && 43.8 & 42.5 \\
         ICR+ReAttn && 77.1 & 85.0 && 59.4 & 56.3 && 89.4 & 94.9 && 43.9 & \bf 42.8 \\
         
        QRhead && 85.5 &  91.7 &&  59.8 & 60.2 &&  93.8 &  96.9 && 47.6 & 41.9 \\
       
        QRhead+ReAttn && \bf 85.8 & \bf 92.4 && \bf 59.9 & \bf 60.5 && \bf 94 & \bf 97.3 && \bf 47.7 & 42 \\

    \bottomrule
    \end{tabular}
    \caption{Performance comparison across different base LLMs and re-ranking methods on LongMemEval and Clipper. The base LLM denotes the LLM used for both the retriever and end-to-end generation.}
    \label{tab:longmemeval}
\end{table*}

\section{Experiments}
\label{sec:main_exp}

\subsection{Experimental Setup}

\paragraph{Datasets.} Following~\cite{zhang2025query}, We evaluate our method on two tasks: re-ranking and long-context reasoning. For re-ranking task, we experiment on eleven public datasets in the BEIR benchmark~\citep{thakur2021beir} consisting of diverse domains, including NQ \citep{NQ}, COVID \citep{voorhees2021treccovid}, NFCorpus \citep{boteva2016nfcorpus}, FiQA \citep{maia2018fiqa},  SciFact \citep{wadden-etal-2020-scifact}, SciDocs \citep{cohan2020scidocs}, FEVER \citep{thorne2018fever}, Climate \citep{diggelmann2020cfever}, DBPedia \citep{hasibi2017dbpedia}, Robust04~\citep{jeronymo2022mrobust04}.  For long-context reasoning task, we use LongMemEval~\citep{wu2025longmemeval} and CLIPPER~\citep{pham2025clipper}. 

% LongMemEval assesses long-term memory in LLM-based dialogue agents, while CLIPPER targets claim verification over book-length texts. Each dataset is segmented according to its inherent structure, using conversational turns in LongMemEval and chapters in CLIPPER. Retrieval performance is measured by recall, and end-to-end performance by accuracy.

\paragraph{Baselines.} 
We compare our approach with several existing re-ranking methods that leverage LLMs, including: (1) RankGPT~\cite{sun2023RankGPT}: a generation-based re-ranking approach that prompts LLMs to generate the ranking order list of a candidate document set conditioned on the query. We evaluate two variants of RankGPT: \textbf{RankGPT\textsuperscript{w/o}}, which feeds all candidate documents into the model prompt simultaneously without sliding window, and \textbf{RankGPT\textsuperscript{Bubble}}, which employs a bubble-sort strategy to iteratively rank smaller subsets of documents. (2) \textbf{ICR}~\citep{chenattention}: an attention-based re-ranking approach that aggregates the attention signals across all heads and layers in LLMs to compute document relevance scores. (3) \textbf{QRhead}~\cite{zhang2025query}: an attention-based re-ranking approach. Instead of aggregating attention weights from all heads and layers, QRhead first identify specialized attention heads that attend selectively to information most relevant to the query on a labeled dataset. Only the attention signals from these selected heads are used to compute relevance scores.

We also include several representative traditional retrievers as baselines, including Contriever, GTR-T5-base, BGE-Reranker-base, MSMARCO-MiniLM and Stella.

\paragraph{Base LLMs.} As attention-based re-ranking method requires access to the  attention weights across all layers and heads of an LLM, we conduct experiments using open-weight LLMs. Specifically, we select three widely used instruction-tuned LLMs from two model families of different scales, including Llama-3.2 (3B), Llama-3.1 (8B) from the Llama family, and Qwen2.5 (7B) from the Qwen family.

\paragraph{Setting.}  Our setting largely follows prior works~\citep{chenattention,zhang2025query}, we re-rank the top 200 documents returned by BM25\citep{bm25} and report nDCG@10. We sub-sampled 512 random questions for each domain for evaluation.  We apply a sliding window of size 20 and stride 10 for RankGPT. For ICR and QRhead, we directly re-rank all documents.

\subsection{Experiment results}

\label{sec:results}
\paragraph{Re-ranking Task.} Table~\ref{tab:beir_main} reports the results on the BEIR benchmark. It demonstrate that ReAttn consistently improves over the baseline attention-based re-rankers, confirming the effectiveness of refining intrinsic attention signals through IDF and entropy regularization. Across all settings, adding ReAttn to ICR or QRhead leads to steady performance gains.  Specifically, under Llama-3.2-3B-Instruct, ICR+ReAttn improves the average nDCG@10 from 46.5 to 47.3, and QRhead+ReAttn further lifts it from 48.6 to 49.3.  A similar trend holds for Llama-3.1-8B-Instruct, where ICR+ReAttn improves by +0.8 points (48.5~$\rightarrow$~49.3), and QRhead+ReAttn achieves the best average score of 50.9.   For Qwen-2.5-7B-Instruct, improvements are smaller but consistent (+1.0 for ICR and +1.1 for QRhead).  
These consistent gains across distinct model families and parameter scales indicate that ReAttn generalizes well, even when underlying LLM architectures differ in tokenizer or attention mechanisms. Notably, ReAttn brings slightly larger relative improvements on smaller model (e.g. 3B). This is because smaller LLMs often exhibit sharper attention distributions and are therefore more susceptible to attention collapse. The entropy-based regularization in ReAttn helps redistribute attention weights, allowing mid-ranked documents to receive more informative gradients and improving ranking metrics such as nDCG@10, which depend on accurate ordering beyond the top few results. 

The improvements are most prominent on datasets with high lexical overlap or entity-centric structure, such as DBPedia, FiQA, and FEVER.  
For example, on DBPedia, ICR+ReAttn improves by +2.1 points under Llama-3B and +0.6 points under Llama-8B, indicating that IDF weighting effectively suppresses distractors sharing repeated entity tokens with the query.  On FiQA, QRhead+ReAttn achieves +1.1 (Llama-3B) and +1.3 (Llama-8B), suggesting that entropy regularization improves ranking discrimination when relevant evidence is distributed across multiple tokens.

Although ReAttn operates without fine-tuning, its re-ranked results are competitive with supervised cross-encoders.   For instance, QRhead+ReAttn (Llama-8B) achieves 50.9 average nDCG@10, surpassing MSMARCO-MiniLM (48.0) and GTR-T5-base (45.2).  
These results indicate that careful post-hoc refinement of intrinsic attention patterns can yield retrieval performance comparable to trained models, while remaining fully parameter-free. Overall, ReAttn consistently enhances the discrimination of attention-based re-rankers.  
The IDF weighting component mitigates the dominance of ubiquitous lexical overlaps, while entropy regularization improves ranking calibration among mid-ranked documents. Together, these mechanisms address the two main weaknesses of ICR—attention concentration and lexical bias—resulting in more balanced and semantically aligned re-ranking behavior across diverse retrieval tasks.

\begin{table*}[t]
    \centering
    \renewcommand{\tabcolsep}{1.2mm}
    \renewcommand{\arraystretch}{1.2}
    \fontsize{8}{8}\selectfont
  
    \begin{tabular}{lcccccccccccc}
    \toprule
& \bf NQ & \bf COVID & \bf NFCorpus & \bf FiQA & \bf Scifact & \bf Scidocs & \bf FEVER & \bf Climate & \bf DBPedia & \bf Robust04 & \bf News & \bf Avg \\

\midrule
& \multicolumn{11}{c}{\textit{Base LLM: Llama-3.2-3B-Instruct }}  \\
\cmidrule{0-12}

ICR & 49.2 & 72.3 & 33.8 & 31.8 & 73.3 & 17.4 & 82.6 & 24.2 & 34.7 & 47.2 & 44.7 & 46.5 \\
ICR+IDF only & 49.6 & 72.3 & 34.2 & 32.3 & 73.8 & 17.3 & 82.7 & 24.2 & 36.0 & 47.3 & 44.9 & 46.8 \\
ICR+Entropy only & 49.4 & 72.4 & 33.9 & 32.0 & 73.6 & 17.4 & 82.9 & 24.4 & 35.5 & 47.6 & 45.3 & 46.8 \\
ICR+ReAttn & 49.8 & 72.5 & 34.4 & 32.5 & 74.1 & 17.5 & 83.0 & 24.5 & 36.8 & 47.7 & 45.4 & 47.3 \\

QRhead & 54.9 & 77.4 & 35.1 & 35.1 & 74.7 & 18.3 & 83.7 & 24.5 & 36.0 & 49.7 & 45.1 & 48.6 \\
QRhead+IDF only & 55.3 & 77.5 & 36.0 & 35.7 & 75.0 & 18.9 & 84.0 & 25.2 & 37.0 & 49.7 & 45.5 & 49.1 \\
QRhead+Entropy only & 55.2 & 77.5 & 35.6 & 35.5 & 75.1 & 19.0 & 84.1 & 25.0 & 36.8 & 49.8 & 45.7 & 49.0 \\
QRhead+ReAttn & \bf 55.6 & \bf 77.6 & \bf 36.6 & \bf 36.2 & \bf 75.3 & \bf 19.5 & \bf 84.3 & \bf 25.6 & \bf 38.0 & \bf 49.8 & \bf 46.0 & \bf 49.3 \\

\midrule
 
& \multicolumn{11}{c}{\textit{Base LLM: Qwen-2.5-7B-Instruct }} \\
\cmidrule{0-12}

ICR & 43.1 & 66.1 & 32.7 & 27.0 & 71.1 & 16.4 & 79.2 & 19.6 & 35.3 & 43.0 & 40.0 & 43.0 \\
ICR+IDF only & 43.8 & 66.4 & 33.5 & 27.4 & 71.7 & 17.2 & 79.5 & 20.1 & 36.2 & 43.1 & 40.4 & 43.6 \\
ICR+Entropy only & 43.5 & 66.7 & 33.1 & 27.7 & 71.5 & 16.9 & 79.7 & 20.4 & 35.8 & 43.4 & 40.7 & 43.6 \\
ICR+ReAttn & 44.0 & 67.0 & 33.9 & 28.1 & \bf 72.2 & \bf 17.8 & 80.0 & 20.9 & 36.6 & 43.6 & 41.2 & 44.0 \\

QRhead & 49.9 & 67.7 & 33.1 & 29.2 & 71.0 & 15.3 & 80.7 & 20.1 & 35.7 & 43.7 & 39.8 & 44.2 \\
QRhead+IDF only & 50.4 & 67.9 & 33.9 & 29.7 & 71.5 & 16.2 & 81.0 & 20.8 & 36.7 & 44.1 & 40.4 & 44.8 \\
QRhead+Entropy only & 50.2 & 68.3 & 33.5 & 29.9 & 71.3 & 15.8 & 81.3 & 20.9 & 36.3 & 44.4 & 40.6 & 44.8 \\
QRhead+ReAttn & \bf 50.7 & 68.6 & \bf 34.2 & \bf 30.3 & 71.9 & 16.6 & \bf 81.5 & \bf 21.3 & \bf 37.4 & 44.8 & 41.0 & \bf 45.3 \\

\bottomrule
    \end{tabular}
    \caption{Performance comparison across different base LLMs and re-ranking methods on BEIR benchmarks evaluated by nDCG@10. \textbf{Bold} indicates the best results.}
    \label{tab:albation_study}
\end{table*}

\paragraph{Long-Context Reasoning Task}
Table~\ref{tab:longmemeval} reports results on the \textbf{LongMemEval} and \textbf{Clipper} benchmarks, which evaluate retrieval and reasoning over extended contexts. The results consistently demonstrate positive gains across both retrieval and end-to-end settings. Under Llama-3.2-3B and Llama-3.1-8B, ReAttn improves recall@k and slightly raises downstream reasoning accuracy for both ICR and QRhead. The trend is stable across datasets and model sizes, indicating that ReAttn’s refinements to the attention signal generalize well to longer input sequences, where attention saturation and token redundancy become more severe. In long-context scenarios, re-rankers must identify sparse but semantically crucial evidence distributed across a large number of retrieved documents. This setting magnifies the limitations of unadjusted attention scores, which often overemphasize lexical overlaps near query tokens and underweight relevant evidence occurring later in the sequence. The IDF-based re-weighting in ReAttn reduces this redundancy by down-weighting query tokens that repeatedly appear across candidate documents, preventing these common tokens from dominating the attention budget. Meanwhile, the entropy regularization ensures that attention is more evenly distributed across informative spans, rather than collapsing onto a few locally similar segments. Together, these refinements improve document-level coverage and facilitate the retrieval of complementary evidence required for multi-hop reasoning.

\subsection{Ablation Study}
\label{sec:ablation}

We further conduct ablations experiments to disentangle the respective contributions of the cross-document IDF re-weighting and the intra-document entropy regularization in ReAttn. Table~\ref{tab:albation_study} demonstrates a systematic dependence on dataset characteristics. On lexically dominated benchmarks (e.g., NQ, NFCorpus, FiQA, DBPedia), the +IDF only variant recovers the majority of ReAttn’s improvements. This is consistent with the fact that relevance in these collections is strongly associated with explicit keyword matching, where rerankers can be disproportionately influenced by surface-level overlap. By reweighting token contributions according to their distinctiveness across the retrieved set, IDF alone effectively suppresses spurious lexical matches and yields substantial ranking gains. In contrast, for corpora featuring longer documents and higher topical heterogeneity (e.g., Robust04, News, Climate), the +Entropy only variant provides comparatively larger benefits. In such settings, the primary failure mode extends beyond lexical bias: attention distributions within long passages often become excessively concentrated on a small number of tokens, which undermines evidence aggregation and degrades robustness. The entropy term counteracts this degeneracy by discouraging overly peaked attention and promoting broader allocation over informative regions, thereby improving coverage of salient content within each document. On Scidocs, applying IDF in isolation can slightly reduce effectiveness, suggesting that distinctiveness-based down-weighting may inadvertently penalize frequent yet domain-informative scientific terminology. Importantly, the full ReAttn (IDF + entropy) restores and surpasses performance, indicating that the two factors target non-overlapping failure modes and are synergistic rather than substitutable.

To further substantiate these findings, we perform a qualitative analysis on FEVER using ICR as the base re-ranker and visualize token-level attention maps (Table~\ref{tbl:kk},~\ref{tbl:jw},~\ref{tbl:fever_appendix_examples}). Removing the IDF component leads to a pronounced shift in scoring behavior: for distractor documents that share surface tokens with the query, ReAttn w/o IDF assigns inflated attention mass to such overlapping terms, which increases document scores despite weak semantic support. By contrast, the full ReAttn attenuates these generic matches via cross-document IDF re-weighting, resulting in attention patterns that concentrate on contextually diagnostic phrases and evidence-bearing spans rather than repeated or high-frequency query tokens. Collectively, these results establish that IDF re-weighting is critical for decoupling lexical overlap from semantic relevance in token-level reranking. By modulating token contributions based on distinctiveness across candidates, ReAttn preserves sharper discrimination between relevant and irrelevant documents, particularly under hard-negative scenarios dominated by verbatim overlap. When combined with entropy regularization that mitigates attention collapse in long passages, the full ReAttn achieves consistently stronger and more stable ranking behavior than either component in isolation.

\section{Conclusion}

We presented ReAttn, a post-hoc re-weighting strategy for attention-based re-ranking with large language models. ReAttn mitigates signal concentration and lexical bias through cross-document IDF weighting and entropy-based regularization, leading to more reliable and balanced attention-based relevance estimation. Extensive experiments across different datasets and  base LLMs confirm that ReAttn consistently improves performance of existing attention-based re-ranking methods.

\section{Limitations}
Our work has several limitations that suggest promising directions for future research.
First, the evaluation of ReAttn is restricted to general-purpose large language models. Although such models provide a representative basis for zero-shot retrieval, a growing number of LLMs have been fine-tuned specifically for information retrieval. These models often exhibit distinct attention dynamics due to task-oriented optimization and domain-adaptive pretraining. Understanding how ReAttn interacts with such IR-specialized models would provide deeper insight into the generality of its attention refinement mechanism.

Second, ReAttn focuses on post-hoc reweighting of attention scores without explicitly examining the contribution of individual attention heads to relevance estimation. Prior work such as QRhead has initiated exploration in this direction, yet its approach to identifying retrieval-relevant heads remains heuristic and coarse-grained. A more systematic analysis of head-level relevance attribution could enhance both the interpretability and the theoretical grounding of attention-based reranking frameworks.

Third, the current experiments are limited to English-language datasets. Since language models may exhibit divergent attention behaviors across linguistic typologies, assessing the cross-lingual robustness of ReAttn is a critical next step. Extending the analysis to multilingual retrieval scenarios would not only test the stability of the proposed adjustments but also inform the design of more universal attention regularization strategies.

\bibliography{latex/custom}
\clearpage
\appendix

\section{Prompt Templates}
We provide prompt templates used in our experiments for ICR,QRhead and our method in Figure \ref{fig:qr-prompt} and rankGPT in Figure \ref{fig:rankgpt-prompt}.

\begin{figure}[h!]
\centering
\begin{tcolorbox}[colframe=black, boxrule=0.8pt]
\small
\ttfamily
\{\texttt{prompt\_prefix}\} Here are some paragraphs:\\

[1] \{Title 1 (if available)\}\\  
\{Paragraph text 1\}\\

[2] \{Title 2 (if available)\}\\
\{Paragraph text 2\}\\

...\\

Please find information that is relevant to the following query in the paragraphs above.\\

Query: \{query\}\{prompt\_suffix\}
\end{tcolorbox}
\caption{Prompt used for ICR, QRhead and our method.}
\label{fig:qr-prompt}
\end{figure}

\begin{figure}[h!]
\centering
\begin{tcolorbox}[colframe=black, boxrule=0.8pt]
\small
\ttfamily
\{\texttt{prefix}\} This is an intelligent assistant that can rank passages based on their relevancy to the query.\\

The following are \{N\} passages, each indicated by a numbered identifier [i]. I can rank them based on their relevance to the query: "\{query\}"\\

[1] \{Title 1 (if available)\}\\  
\{Paragraph text 1\}\\

[2] \{Title 2 (if available)\}\\
\{Paragraph text 2\}\\

...\\

The search query is: "\{query\}". I will rank the \{N\} passages above based on their relevance to the search query. The passages will be listed in descending order using identifiers, the most relevant passages should be listed first and the output format should be [] > [] > etc, e.g., [1] > [2] > etc. Be sure to list all \{N\} ranked passages and do not explain your ranking until after the list is done. \{suffix\} Ranked Passages: [
\end{tcolorbox}
\caption{Prompt used for rankGPT.}
\label{fig:rankgpt-prompt}
\end{figure}

\setlength{\tabcolsep}{2pt}
\begin{table*}[ht]
  \small
  \centering
  \caption{Attention scores for all tokens in a FEVER example with the query ``Night of the Living Dead was originally directed by Krzysztof Kieslowski.''}
  \begin{tabular}{cp{0.7\linewidth}cc}
  \toprule
   \textbf{Method} & \textbf{Passage} & \textbf{Rank} \\ \midrule
\textsc{ReAttn w/o IDF} &\hlc[cyan!8!white]{The} \hlc[cyan!7!white]{Scar} \hlc[red!0!white]{(} \hlc[cyan!2!white]{Blizna} \hlc[cyan!2!white]{)} \hlc[red!0!white]{is} \hlc[red!2!white]{a} \hlc[cyan!10!white]{1976} \hlc[cyan!22!white]{Polish} \hlc[cyan!7!white]{film} \hlc[cyan!4!white]{written} \hlc[cyan!3!white]{and} \hlc[cyan!6!white]{directed} \hlc[cyan!18!white]{by} \hlc[cyan!48!white]{Krzysztof} \hlc[cyan!89!white]{Kie}ś\hlc[cyan!100!white]{lowski} \hlc[cyan!48!white]{and} \hlc[cyan!13!white]{starring} \hlc[cyan!7!white]{Franciszek} \hlc[cyan!5!white]{Pieczka} \hlc[cyan!6!white]{.} \hlc[cyan!1!white]{Filmed} \hlc[red!1!white]{on} \hlc[red!0!white]{location} \hlc[red!1!white]{in} \hlc[cyan!0!white]{Olech}ó\hlc[cyan!3!white]{w} \hlc[red!0!white]{,} \hlc[cyan!3!white]{Poland} \hlc[red!0!white]{,} \hlc[red!0!white]{the} \hlc[cyan!4!white]{film} \hlc[red!0!white]{is} \hlc[cyan!0!white]{about} \hlc[cyan!1!white]{a} \hlc[cyan!2!white]{man} \hlc[cyan!1!white]{put} \hlc[cyan!1!white]{in} \hlc[cyan!4!white]{charge} \hlc[cyan!1!white]{of} \hlc[cyan!1!white]{the} \hlc[cyan!3!white]{construction} ...  & 3\\
\midrule
\textsc{ReAttn}&\hlc[cyan!8!white]{The} \hlc[cyan!7!white]{Scar} \hlc[red!0!white]{(} \hlc[cyan!2!white]{Blizna} \hlc[cyan!2!white]{)} \hlc[red!0!white]{is} \hlc[red!2!white]{a} \hlc[cyan!10!white]{1976} \hlc[cyan!22!white]{Polish} \hlc[cyan!7!white]{film} \hlc[cyan!4!white]{written} \hlc[cyan!3!white]{and} \hlc[cyan!6!white]{directed} \hlc[cyan!18!white]{by} \hlc[cyan!30!white]{Krzysztof} \hlc[cyan!20!white]{Kie}ś\hlc[cyan!20!white]{lowski} \hlc[cyan!48!white]{and} \hlc[cyan!13!white]{starring} \hlc[cyan!7!white]{Franciszek} \hlc[cyan!5!white]{Pieczka} \hlc[cyan!6!white]{.} \hlc[cyan!1!white]{Filmed} \hlc[red!1!white]{on} \hlc[red!0!white]{location} \hlc[red!1!white]{in} \hlc[cyan!0!white]{Olech}ó\hlc[cyan!3!white]{w} \hlc[red!0!white]{,} \hlc[cyan!3!white]{Poland} \hlc[red!0!white]{,} \hlc[red!0!white]{the} \hlc[cyan!4!white]{film} \hlc[red!0!white]{is} \hlc[cyan!0!white]{about} \hlc[cyan!1!white]{a} \hlc[cyan!2!white]{man} \hlc[cyan!1!white]{put} \hlc[cyan!1!white]{in} \hlc[cyan!4!white]{charge} \hlc[cyan!1!white]{of} \hlc[cyan!1!white]{the} \hlc[cyan!3!white]{construction} ... & 6  \\
  \bottomrule
  \end{tabular}%
  \label{tbl:kk}
\end{table*}

\begin{table*}[ht]
  \small
  \centering
  \caption{Attention scores for all tokens in a FEVER example with the query ``Jiang Wen is exclusively a producer.''}
  \begin{tabular}{cp{0.7\linewidth}c}
  \toprule
  \textbf{Method} & \textbf{Passage}  & \textbf{Rank} \\ \midrule
\textsc{ReAttn w/o IDF} & \hlc[cyan!31!white]{Emperor} \hlc[cyan!19!white]{Motion} \hlc[cyan!16!white]{Pictures} \hlc[cyan!4!white]{(} \hlc[cyan!3!white]{known} \hlc[cyan!3!white]{as} \hlc[cyan!32!white]{EMP} \hlc[cyan!5!white]{)} \hlc[cyan!9!white]{is} \hlc[cyan!9!white]{a} \hlc[cyan!33!white]{film} \hlc[cyan!100!white]{producer} \hlc[cyan!37!white]{and} \hlc[cyan!44!white]{distributor} \hlc[cyan!20!white]{,} \hlc[cyan!5!white]{part} \hlc[cyan!5!white]{of} \hlc[cyan!1!white]{the} \hlc[cyan!7!white]{Emperor} \hlc[cyan!9!white]{Group} \hlc[cyan!10!white]{.} \hlc[cyan!3!white]{Following} \hlc[cyan!0!white]{the} \hlc[cyan!2!white]{2003} \hlc[cyan!3!white]{box}-\hlc[cyan!1!white]{office} \hlc[cyan!2!white]{hits} \hlc[cyan!1!white]{The} \hlc[cyan!1!white]{Twins} \hlc[cyan!2!white]{Effect} \hlc[cyan!3!white]{and} \hlc[cyan!0!white]{The} \hlc[cyan!0!white]{Medallion} \hlc[cyan!4!white]{,} \hlc[cyan!34!white]{EMP} \hlc[cyan!5!white]{has} \hlc[cyan!45!white]{produced} ... \hlc[cyan!0!white]{The} \hlc[cyan!1!white]{Sun} \hlc[cyan!11!white]{Also} \hlc[cyan!30!white]{Rises} \hlc[cyan!6!white]{and} \hlc[cyan!1!white]{Forever} \hlc[cyan!0!white]{Enthralled} \hlc[cyan!0!white]{,} \hlc[cyan!1!white]{two} \hlc[cyan!4!white]{works} \hlc[cyan!8!white]{by} \hlc[cyan!4!white]{renowned} \hlc[cyan!19!white]{Chinese} \hlc[cyan!16!white]{auteurs} \hlc[cyan!78!white]{Jiang} \hlc[cyan!66!white]{Wen} \hlc[cyan!59!white]{and} \hlc[cyan!5!white]{Chen} \hlc[cyan!3!white]{Kaige}\hlc[cyan!40!white]{.} ...  & 2 \\
\midrule
\textsc{ReAttn}& \hlc[cyan!31!white]{Emperor} \hlc[cyan!19!white]{Motion} \hlc[cyan!16!white]{Pictures} \hlc[cyan!4!white]{(} \hlc[cyan!3!white]{known} \hlc[cyan!3!white]{as} \hlc[cyan!32!white]{EMP} \hlc[cyan!5!white]{)} \hlc[cyan!9!white]{is} \hlc[cyan!9!white]{a} \hlc[cyan!33!white]{film} \hlc[cyan!20!white]{producer} \hlc[cyan!37!white]{and} \hlc[cyan!44!white]{distributor} \hlc[cyan!20!white]{,} \hlc[cyan!5!white]{part} \hlc[cyan!5!white]{of} \hlc[cyan!1!white]{the} \hlc[cyan!7!white]{Emperor} \hlc[cyan!9!white]{Group} \hlc[cyan!10!white]{.} \hlc[cyan!3!white]{Following} \hlc[cyan!0!white]{the} \hlc[cyan!2!white]{2003} \hlc[cyan!3!white]{box}-\hlc[cyan!1!white]{office} \hlc[cyan!2!white]{hits} \hlc[cyan!1!white]{The} \hlc[cyan!1!white]{Twins} \hlc[cyan!2!white]{Effect} \hlc[cyan!3!white]{and} \hlc[cyan!0!white]{The} \hlc[cyan!0!white]{Medallion} \hlc[cyan!4!white]{,} \hlc[cyan!34!white]{EMP} \hlc[cyan!5!white]{has} \hlc[cyan!45!white]{produced} ... \hlc[cyan!0!white]{The} \hlc[cyan!1!white]{Sun} \hlc[cyan!11!white]{Also} \hlc[cyan!30!white]{Rises} \hlc[cyan!6!white]{and} \hlc[cyan!1!white]{Forever} \hlc[cyan!0!white]{Enthralled} \hlc[cyan!0!white]{,} \hlc[cyan!1!white]{two} \hlc[cyan!4!white]{works} \hlc[cyan!8!white]{by} \hlc[cyan!4!white]{renowned} \hlc[cyan!19!white]{Chinese} \hlc[cyan!16!white]{auteurs} \hlc[cyan!20!white]{Jiang} \hlc[cyan!20!white]{Wen} \hlc[cyan!59!white]{and} \hlc[cyan!5!white]{Chen} \hlc[cyan!3!white]{Kaige}\hlc[cyan!40!white]{.} ... & 10 \\
  \bottomrule
  \end{tabular}%
  \label{tbl:jw}
\end{table*}

\setlength{\tabcolsep}{2pt}
\begin{table*}[h]
  \small
  \centering
  \caption{Attention scores for all tokens in a DBPedia-Entity example with the query ``Give me all launch pads operated by NASA.'' }
  \begin{tabular}{cp{0.7\linewidth}c}
  \toprule
  \textbf{Method} & \textbf{Passage} &\textbf{Rank} \\ \midrule
 \textsc{ReAttn w/o IDF}&  \hlc[cyan!0!white]{Cape} \hlc[cyan!5!white]{Can} \hlc[cyan!23!white]{av} \hlc[cyan!7!white]{eral} \hlc[cyan!3!white]{Air} \hlc[cyan!3!white]{Force} \hlc[cyan!7!white]{Station} \hlc[cyan!12!white]{Launch} \hlc[cyan!8!white]{Complex} \hlc[cyan!6!white]{} \hlc[cyan!12!white]{19} \hlc[cyan!5!white]{Ċ} \hlc[cyan!15!white]{Launch} \hlc[cyan!4!white]{Complex} \hlc[cyan!2!white]{} \hlc[cyan!1!white]{19} \hlc[cyan!1!white]{(} \hlc[cyan!1!white]{LC} \hlc[cyan!4!white]{-} \hlc[cyan!3!white]{19} \hlc[cyan!2!white]{)} \hlc[cyan!1!white]{is} \hlc[cyan!1!white]{a} \hlc[cyan!14!white]{deactivated} \hlc[cyan!10!white]{launch} \hlc[cyan!9!white]{site} \hlc[cyan!4!white]{on} \hlc[cyan!3!white]{Cape} \hlc[cyan!4!white]{Can} \hlc[red!0!white]{av} \hlc[cyan!2!white]{eral} \hlc[cyan!3!white]{Air} \hlc[cyan!1!white]{Force} \hlc[cyan!2!white]{Station} \hlc[cyan!2!white]{,} \hlc[cyan!3!white]{Florida} \hlc[cyan!0!white]{} \hlc[cyan!2!white]{used} \hlc[cyan!4!white]{by} \hlc[cyan!60!white]{NASA} \hlc[cyan!8!white]{to} \hlc[cyan!20!white]{launch} \hlc[cyan!4!white]{all} \hlc[cyan!2!white]{of} \hlc[cyan!1!white]{the} \hlc[cyan!2!white]{Gemini} \hlc[cyan!3!white]{manned} \hlc[cyan!2!white]{space} \hlc[cyan!5!white]{fl} \hlc[cyan!4!white]{ights} \hlc[cyan!4!white]{.} \hlc[cyan!2!white]{It} \hlc[cyan!1!white]{was} \hlc[cyan!1!white]{also} \hlc[cyan!2!white]{used} \hlc[cyan!2!white]{by} \hlc[cyan!4!white]{unmanned} \hlc[cyan!1!white]{Titan} \hlc[cyan!1!white]{I} \hlc[cyan!1!white]{and} \hlc[cyan!0!white]{Titan} \hlc[cyan!0!white]{II} \hlc[cyan!2!white]{missiles} \hlc[red!0!white]{.L} \hlc[cyan!0!white]{C} \hlc[cyan!2!white]{-} \hlc[cyan!1!white]{19} \hlc[cyan!0!white]{was} \hlc[cyan!1!white]{in} \hlc[cyan!3!white]{use} \hlc[cyan!1!white]{from} \hlc[cyan!0!white]{} \hlc[cyan!2!white]{195} \hlc[cyan!1!white]{9} \hlc[cyan!0!white]{to} \hlc[cyan!0!white]{} \hlc[cyan!0!white]{196} \hlc[cyan!0!white]{6} \hlc[cyan!1!white]{,} \hlc[cyan!0!white]{during} \hlc[cyan!0!white]{which} \hlc[cyan!0!white]{time} \hlc[cyan!0!white]{it} \hlc[cyan!0!white]{saw} \hlc[cyan!1!white]{} \hlc[red!1!white]{27} \hlc[cyan!10!white]{launches} \hlc[cyan!2!white]{,} \hlc[cyan!0!white]{} \hlc[red!0!white]{10} \hlc[red!0!white]{of} \hlc[cyan!0!white]{which} \hlc[cyan!0!white]{were} \hlc[cyan!0!white]{manned} \hlc[cyan!0!white]{.} \hlc[cyan!0!white]{The} \hlc[red!0!white]{first} \hlc[cyan!0!white]{use} \hlc[cyan!0!white]{of} \hlc[cyan!2!white]{LC} \hlc[cyan!7!white]{-} \hlc[cyan!0!white]{19} \hlc[cyan!0!white]{was} \hlc[red!0!white]{on} \hlc[red!0!white]{August} \hlc[red!0!white]{} \hlc[red!0!white]{14} \hlc[cyan!2!white]{,} \hlc[cyan!0!white]{} \hlc[cyan!0!white]{195} \hlc[red!0!white]{9} \hlc[cyan!0!white]{.} \hlc[red!0!white]{This} \hlc[red!0!white]{was} \hlc[red!0!white]{a} \hlc[red!0!white]{Titan} \hlc[cyan!0!white]{I} \hlc[cyan!0!white]{and} \hlc[cyan!0!white]{the} \hlc[cyan!0!white]{mission} \hlc[cyan!0!white]{was} \hlc[red!0!white]{declared} \hlc[cyan!0!white]{a} \hlc[cyan!0!white]{failure} \hlc[red!0!white]{after} \hlc[red!0!white]{the} \hlc[cyan!1!white]{rocket} \hlc[cyan!0!white]{exploded} \hlc[red!0!white]{while} \hlc[red!0!white]{still} \hlc[cyan!0!white]{on} \hlc[cyan!0!white]{the} \hlc[cyan!18!white]{pad} \hlc[cyan!0!white]{.}  & 12  \\
 \midrule

\textsc{ReAttn}& \hlc[cyan!0!white]{Cape} \hlc[cyan!5!white]{Can} \hlc[cyan!23!white]{av} \hlc[cyan!7!white]{eral} \hlc[cyan!3!white]{Air} \hlc[cyan!3!white]{Force} \hlc[cyan!7!white]{Station} \hlc[cyan!7!white]{Launch} \hlc[cyan!8!white]{Complex} \hlc[cyan!6!white]{} \hlc[cyan!12!white]{19} \hlc[cyan!5!white]{Ċ} \hlc[cyan!1!white]{Launch} \hlc[cyan!4!white]{Complex} \hlc[cyan!2!white]{} \hlc[cyan!1!white]{19} \hlc[cyan!1!white]{(} \hlc[cyan!1!white]{LC} \hlc[cyan!4!white]{-} \hlc[cyan!3!white]{19} \hlc[cyan!2!white]{)} \hlc[cyan!1!white]{is} \hlc[cyan!1!white]{a} \hlc[cyan!14!white]{deactivated} \hlc[cyan!7!white]{launch} \hlc[cyan!9!white]{site} \hlc[cyan!4!white]{on} \hlc[cyan!3!white]{Cape} \hlc[cyan!4!white]{Can} \hlc[red!0!white]{av} \hlc[cyan!2!white]{eral} \hlc[cyan!3!white]{Air} \hlc[cyan!1!white]{Force} \hlc[cyan!2!white]{Station} \hlc[cyan!2!white]{,} \hlc[cyan!3!white]{Florida} \hlc[cyan!0!white]{} \hlc[cyan!2!white]{used} \hlc[cyan!4!white]{by} \hlc[cyan!20!white]{NASA} \hlc[cyan!8!white]{to} \hlc[cyan!4!white]{launch} \hlc[cyan!4!white]{all} \hlc[cyan!2!white]{of} \hlc[cyan!1!white]{the} \hlc[cyan!2!white]{Gemini} \hlc[cyan!3!white]{manned} \hlc[cyan!2!white]{space} \hlc[cyan!5!white]{fl} \hlc[cyan!4!white]{ights} \hlc[cyan!4!white]{.} \hlc[cyan!2!white]{It} \hlc[cyan!1!white]{was} \hlc[cyan!1!white]{also} \hlc[cyan!2!white]{used} \hlc[cyan!2!white]{by} \hlc[cyan!4!white]{unmanned} \hlc[cyan!1!white]{Titan} \hlc[cyan!1!white]{I} \hlc[cyan!1!white]{and} \hlc[cyan!0!white]{Titan} \hlc[cyan!0!white]{II} \hlc[cyan!2!white]{missiles} \hlc[red!0!white]{.L} \hlc[cyan!0!white]{C} \hlc[cyan!2!white]{-} \hlc[cyan!1!white]{19} \hlc[cyan!0!white]{was} \hlc[cyan!1!white]{in} \hlc[cyan!3!white]{use} \hlc[cyan!1!white]{from} \hlc[cyan!0!white]{} \hlc[cyan!2!white]{195} \hlc[cyan!1!white]{9} \hlc[cyan!0!white]{to} \hlc[cyan!0!white]{} \hlc[cyan!0!white]{196} \hlc[cyan!0!white]{6} \hlc[cyan!1!white]{,} \hlc[cyan!0!white]{during} \hlc[cyan!0!white]{which} \hlc[cyan!0!white]{time} \hlc[cyan!0!white]{it} \hlc[cyan!0!white]{saw} \hlc[cyan!1!white]{} \hlc[red!1!white]{27} \hlc[cyan!5!white]{launches} \hlc[cyan!2!white]{,} \hlc[cyan!0!white]{} \hlc[red!0!white]{10} \hlc[red!0!white]{of} \hlc[cyan!0!white]{which} \hlc[cyan!0!white]{were} \hlc[cyan!0!white]{manned} \hlc[cyan!0!white]{.} \hlc[cyan!0!white]{The} \hlc[red!0!white]{first} \hlc[cyan!0!white]{use} \hlc[cyan!0!white]{of} \hlc[cyan!2!white]{LC} \hlc[cyan!7!white]{-} \hlc[cyan!0!white]{19} \hlc[cyan!0!white]{was} \hlc[red!0!white]{on} \hlc[red!0!white]{August} \hlc[red!0!white]{} \hlc[red!0!white]{14} \hlc[cyan!2!white]{,} \hlc[cyan!0!white]{} \hlc[cyan!0!white]{195} \hlc[red!0!white]{9} \hlc[cyan!0!white]{.} \hlc[red!0!white]{This} \hlc[red!0!white]{was} \hlc[red!0!white]{a} \hlc[red!0!white]{Titan} \hlc[cyan!0!white]{I} \hlc[cyan!0!white]{and} \hlc[cyan!0!white]{the} \hlc[cyan!0!white]{mission} \hlc[cyan!0!white]{was} \hlc[red!0!white]{declared} \hlc[cyan!0!white]{a} \hlc[cyan!0!white]{failure} \hlc[red!0!white]{after} \hlc[red!0!white]{the} \hlc[cyan!1!white]{rocket} \hlc[cyan!0!white]{exploded} \hlc[red!0!white]{while} \hlc[red!0!white]{still} \hlc[cyan!0!white]{on} \hlc[cyan!0!white]{the} \hlc[cyan!7!white]{pad} \hlc[cyan!0!white]{.}  & 18  \\

  \bottomrule
  \end{tabular}%
  \label{tbl:fever_appendix_examples}
\end{table*}

\end{document}